\documentclass[conference]{IEEEtran}
\IEEEoverridecommandlockouts
\usepackage{cite}
\usepackage{amsmath,amssymb,amsfonts}
\usepackage{algorithmic}
\usepackage{graphicx}
\usepackage{textcomp}
\usepackage{xcolor}

\usepackage{times}
\usepackage{latexsym}
\usepackage{graphicx} 
\usepackage{subcaption}
\usepackage{tikz}
\usepackage{multirow}
\usepackage{xspace}
\usepackage{tabularray}
\usepackage{booktabs}

\usepackage{inconsolata}
\usepackage{enumitem}
\usepackage{hyperref}
\usepackage{subcaption}

\newcommand{\todo}[1]{{\color{red}[#1]}}
\newcommand{\ignore}[1]{}
\newcommand{\ignorebecauseofproduct}[1]{}
\newcommand{\predicate}[1]{{\tt #1}}
\newcommand{\triple}[1]{{\tt #1}}
\newcommand{\spotriple}[1]{$\langle${\tt#1}$\rangle$}

\newcommand{\fandom}{{\tt platform-1}\xspace}
\newcommand{\famousbirthday}{{\tt platform-2}\xspace}
\newcommand{\system}{ODKE\xspace}
\newcommand{\systemWikipedia}{ODKE-Wikipedia\xspace}

\def\BibTeX{{\rm B\kern-.05em{\sc i\kern-.025em b}\kern-.08em
    T\kern-.1667em\lower.7ex\hbox{E}\kern-.125emX}}
\begin{document}

\title{Open Domain Knowledge Extraction for Knowledge Graphs}

\author{\IEEEauthorblockN{Kun Qian, Anton Belyi, Fei Wu, Samira Khorshidi, Azadeh Nikfarjam, \\
Rahul Khot, Yisi Sang, Katherine Luna, Xianqi Chu, Eric Choi, \\
Yash Govind, Chloe Seivwright, Yiwen Sun, Ahmed Fakhry, \\
Theo Rekatsinas, Ihab Ilyas, Xiaoguang Qi, Yunyao Li
}
\IEEEauthorblockA{\textit{Apple} \\
\{kunqian, a\_belyy, fwu7, samiraa, anikfarjam\}@apple.com\\
\{r\_khot, yisi\_sang, kluna, xchu23, eyhchoi\}@apple.com\\
\{yash\_govind, cseivwright, yiwen\_sun, afakhry\}@apple.com\\
\{trekatsinas, iilyas, xiaoguang\_qi, yunyaoli\}@apple.com
}
}

\maketitle

\begin{abstract}
The quality of a knowledge graph directly impacts the quality of downstream applications (e.g. the number of answerable questions using the graph). One ongoing challenge when building a knowledge graph is to ensure completeness and freshness of the graph's entities and facts. In this paper, we introduce \system, a scalable and extensible framework that sources high-quality entities and facts from open web at scale. 
\system utilizes a wide range of extraction models and supports both streaming and batch processing at different latency. We reflect on the challenges and design decisions made and share lessons learned when building and deploying \system to grow an industry-scale open domain knowledge graph. 
\end{abstract}


\section{Introduction}
A knowledge graph (KG) organizes open-domain knowledge in a structured way by capturing relationships and semantic connections between entities. It provides a comprehensive and interconnected view of information that powers many real-world applications, such as question answering, relationship extraction, entity disambiguation, and data integration \cite{saga_sigmod, huang2019knowledge}. The usefulness and importance of knowledge graphs become even more pronounced in the era of large language models (LLMs), as LLMs are known for lack of factual knowledge and therefore often hallucinate factually incorrect claims \cite{pan2023unifying, survey_hallucination2023}. Curated KGs, known for their high quality and reliability, offer dependable, structured, and explainable knowledge, which black-box models like LLMs are unable to offer \cite{kg-survey, never-ending-learning}. 

 Traditionally, ensuring accurate knowledge ingestion into KGs has involved tedious and costly human curation \cite{kamath-etal-2022-improving}. To avoid or mitigate the labor intensive and obviously non-scalable process, it is essential to develop an automated knowledge extraction and ingestion framework that can continuously update KGs with highly accurate facts to maintain their completeness and freshness. However, several challenges must be addressed in order to achieve this goal:
\begin{figure*}[ht!]
\centering
\valign{#\cr
  \hbox{%
    \begin{subfigure}[b]{.2\textwidth}
    \centering
    \includegraphics[height=5.9cm, width=\textwidth]{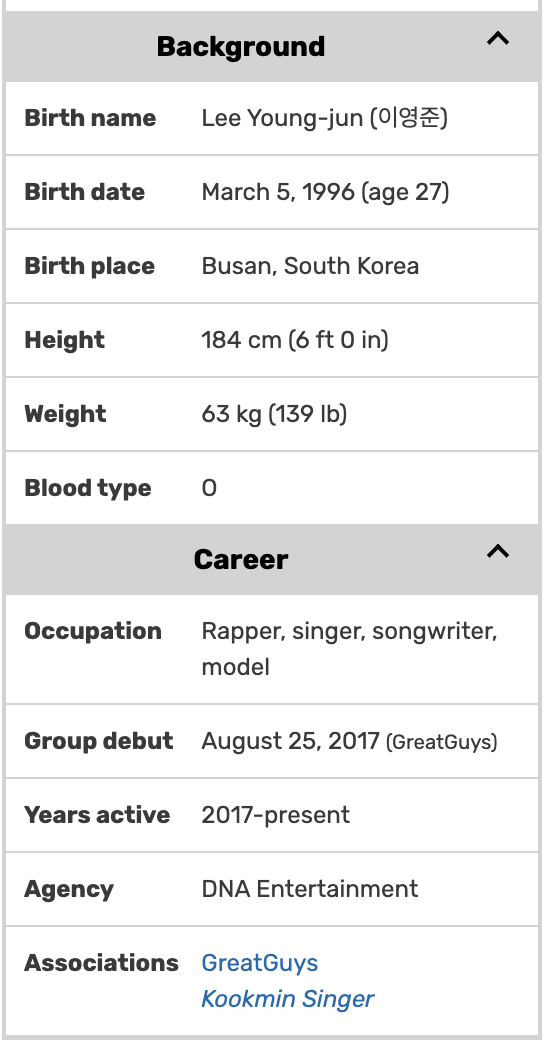}
    \caption{}
    \end{subfigure}%
  }
  \cr\noalign{\hfill}
  \hbox{%
    \begin{subfigure}[b]{.48\textwidth}
    \centering
    \fbox{\includegraphics[width=\textwidth]{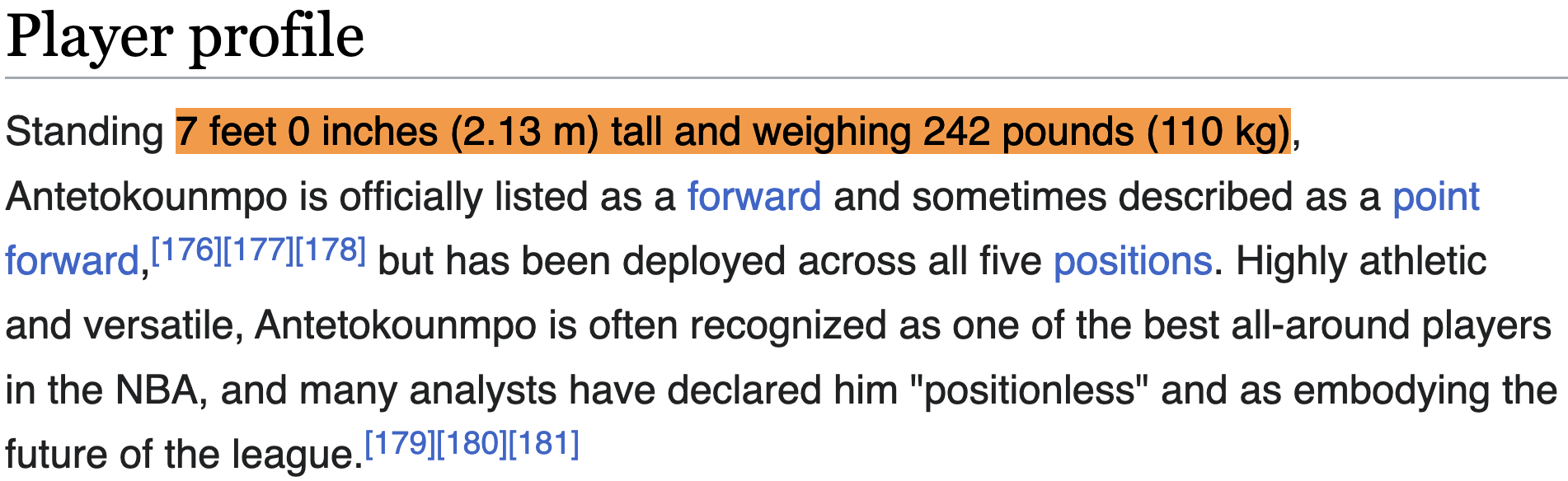}}
    \caption{}
    \end{subfigure}%
  }\vfill
  \hbox{%
    \begin{subfigure}[b]{.48\textwidth}
    \centering
    \fbox{\includegraphics[width=\textwidth]{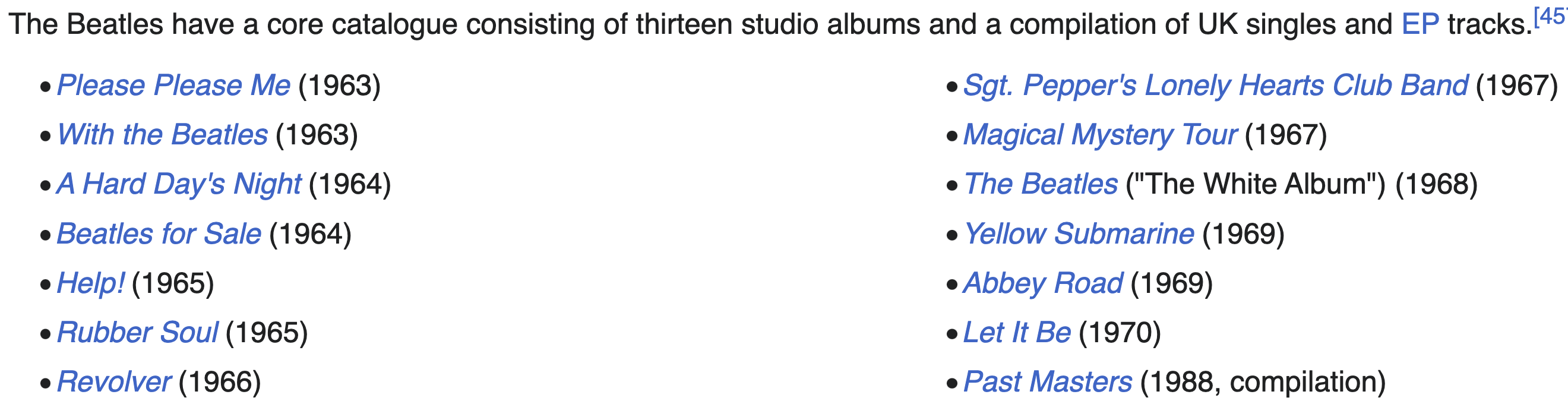}}
    \caption{}
    \end{subfigure}%
  }
  \cr
  \noalign{\hfill}
  \hbox{%
    \begin{subfigure}{.22\textwidth}
    \centering
    \fbox{\includegraphics[width=\textwidth]{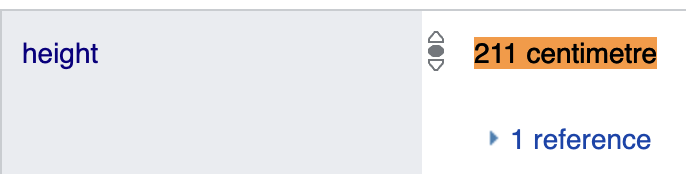}}
    \caption{}
    \end{subfigure}%
  }\vfill
  \hbox{%
    \begin{subfigure}{.22\textwidth}
    \centering
    \fbox{\includegraphics[width=\textwidth]{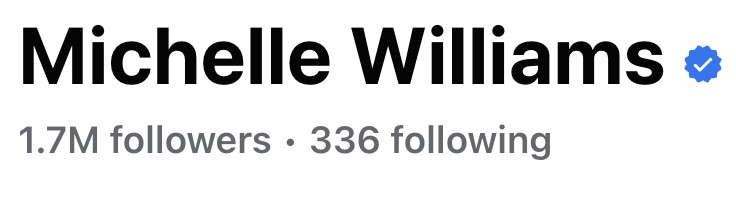}}
    \caption{}
    \end{subfigure}%
  }
  \hbox{%
    \begin{subfigure}{.22\textwidth}
    \centering
    \fbox{\includegraphics[width=\textwidth]{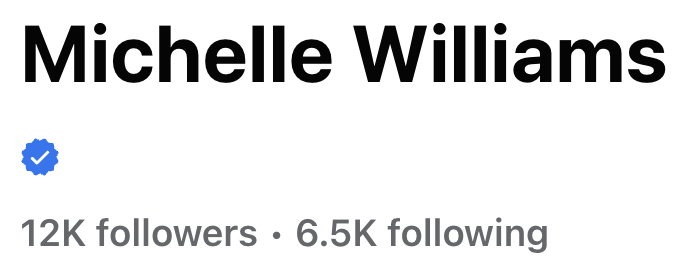}}
    \caption{}
    \end{subfigure}%
  }\cr
}
\caption{Examples of different data modalities for knowledge extraction: (a) Web table from fandom.com; (b) and (d) are the height values of NBA player Antetokounmpo from Wikpedia (as plain text) and from Wikidata as a semi-structured key-value pair; (c) Discography of The Beatles from a web list. Ambiguous entities: (e) and (f) are two verified Facebook pages for two entities who share the same name {\em Michelle Williams}.}
\label{fig:example}
\end{figure*}

\begin{itemize} 
\item {\textbf{Large volume of data}}. The amount of data and facts contained on the Web is enormous and continuously growing. According to an estimation by \texttt{\href{https://WorldWideWebSize.com}{WorldWideWebSize.com}}, there are about 40-50 billion web pages indexed by Google as of July 2023. In fact, Wikipedia alone contains ~58M articles \cite{Wikipedia_size}.  We need to handle the scalability challenge posed by Web-scale data.  
\smallskip
\item {\textbf{Wide variety of data and tasks}}. The Web contains a wide variety of data, from plain text to semi-structured data and mixtures of both. Figure \ref{fig:example} illustrates the various forms in which knowledge can be expressed on the open Web.
To extract high-quality facts from the Web, we need extractors that are capable of extracting high-quality facts for different types of entities in various modalities from different data sources. 
\smallskip
\item {\textbf{High veracity}}. The Web is noisy and often contains wrong and conflicting facts, e.g., NBA player Antetokounmpo's height is {\em ``2.13 m''} according to his Wikipedia page (Fig. \ref{fig:example}(b)) and \href{https://www.nba.com/player/203507/giannis-antetokounmpo}{NBA official page} (accessed June 2023), while his \href{https://www.Wikidata.org/wiki/Q8991894}{Wikidata page} shows {\em ``211 cm''} (Fig.\ref{fig:example}(d)). Additionally, certain facts, such as an individual's marital status and the head coach of a professional sports team, may change over time. As such, it becomes imperative to identify the most accurate and current facts.
\smallskip
\item {\textbf{Various velocity}}. Timely extracting fresh knowledge from Web and ingesting into KG is crucial for many downstream applications.  Everyday, new popular entities such as trendy YouTubers and TikTokers emerge, and their fans eagerly seek up-to-date information about them. Take English Wikipedia pages as an example, roughly 2 million English Wikipedia pages got edited monthly and an average of 549 new articles are created per day \cite{wikimedia_en_Wikipedia_stats}. Although non-time-sensitive facts can be updated in a weekly batch fashion, some facts need to be updated much more faster (e.g., newly announced Academy Award winners). 
\end{itemize}

Most prior work focuses on some individual problems such as extracting knowledge from semi-structured data \cite{opentag, knowledge-vault, knowledge-fusion}. 
Few of these works actually build an end-to-end automated extraction and ingestion framework at industrial scale. One exception is \cite{kb-completion-search-based}, which proposed an end-to-end KG completion framework with search-based question answering. The core idea is to learn the set of queries to ask and retrieve answers from a search engine followed by a simple ranking-based scoring approach to find the best answer to complete a missing fact in the KG. While this solution is reasonable, it is limited by the potential incomplete search engine results, limited coverage of specialized domains or knowledge, and inability to incorporate more advanced techniques such as LLM-based extraction.  

To fully address the aforementioned challenges at scale, we propose \system, an automated and powerful knowledge extraction and ingestion framework for growing the coverage and freshness of an industry-scale open domain KG. The initial version of ODKE was first introduced in \cite{Ilyas_2023} as the extraction component of the Saga system \cite{saga}, and in this paper, we describe the newer version of ODKE that greatly extends the functionalities and capabilities of the framework described in \cite{Ilyas_2023}. Table \ref{tab:odke-vs-Wikidata} summarizes the major extensions of the present ODKE framework (i.e., ODKE v2, the present paper) over the previous ODKE framework (i.e., ODKE v1 \cite{Ilyas_2023}). 
Concretely, \system v1 adopts a solution that is very similar to \cite{kb-completion-search-based}, which mainly focus on reactively trigger extraction through missing fact identification and use search engines to collect candidate documents for extraction on the fly even in batch extraction mode. In \system v2, we extend the framework so that we can identify some highly important data sources such as Wikipedia, and we can continuously crawl and monitor the changes from upstream data sources and trigger either batch extraction or continuous extraction (\ref{sec:streaming}) with \system depending on the applications. This greatly improved the scalability and freshness of our extraction pipelines. Another major extension is multilingual support and link inference. We are carefully redesigned our knowledge extractors, where we utilizes language-specific patterns and langauage-agnostic LLMs so that we can support extracting facts in different languages. Besides the main extraction framework, we also add a new component called Link Inference (\ref{sec:linkinference}), which add missing links in KG without perform actual extractions. More details will be provided in next sections of the paper.

\begin{table}[]
\centering
\normalsize
\renewcommand{\arraystretch}{1.2}
\begin{tabular}{lcc}
\multicolumn{1}{c}{}            & \textbf{ODKE v1 \cite{Ilyas_2023}}   & \textbf{ODKE v2}                                                   \\
\specialrule{\heavyrulewidth}{0pt}{\belowrulesep}
\textbf{Evidence Retrieval}     & Search-based       & \begin{tabular}[c]{@{}c@{}}Search-based\\ Crawl-based\end{tabular} \\
\cmidrule(lr){2-3} 
\textbf{Extraction Power}       & Pattern-based      & \begin{tabular}[c]{@{}c@{}}Pattern-based\\ LLM-based\end{tabular}  \\
\cmidrule(lr){2-3}  
\textbf{Multilingual Support}   & No                 & Yes  \\
\cmidrule(lr){2-3} 
\textbf{Link Inference Support} & No                 & Yes                                                                \\
\cmidrule(lr){2-3}
\textbf{Streaming Support} & No                 & Yes                                                                \\

\cmidrule(lr){2-3} 
\textbf{Stability}              & \begin{tabular}[c]{@{}c@{}}up to\\ 5k facts/min\end{tabular} & \begin{tabular}[c]{@{}c@{}}up to\\ 100k facts/min\end{tabular} \\
\specialrule{\heavyrulewidth}{0pt}{\belowrulesep}
\end{tabular}
\caption{Comparison of ODKE v1 \cite{Ilyas_2023} and ODKE v2}
\label{table:odke-v1-v2}
\end{table}

\section{System Architecture}
\label{sec:arch}
Fig. \ref{fig:architecture} shows the extraction and ingestion pipelines supported by \system. 
 The process begins with the Extraction Initiator, which identifies missing or stale facts in the KG or specifies a web data source for extraction.
The Retriever collects web documents (e.g., through open web search and/or web crawling) for extraction. Various type of extractors are supported by \system, including pattern-based, traditional ML-based, and LLM-based extractors.
Corroborator normalizes and clusters the facts, which are then scored and re-ranked by Scorer.
High-confidence facts are automatically exported and ingested into the KG, while sensitive and uncertain extractions will be scrutinized by human curators. 
Moreover, \system periodically performs link inference on the KG to add missing links without actual extraction.
Each component of \system is described in detail in the following subsections.

\begin{figure*}
\centering
\input{figures/latex-figure/architecture}
\caption{\system Architecture}
\label{fig:architecture}
\end{figure*}

\subsection{Extraction Initiator}
\label{sec:initiator}
The Extraction Initiator is a module that determines what information \system needs to retrieve and standardize into a common format for the downstream application. 
To trigger extraction in \system, we can either reactively identify stale and missing facts through KG profiling or user feedback, or we can identify reliable knowledge data sources (e.g., Wikipedia) and extract facts all facts that are supported by our knowledge extractors. 

We aggregate different sources of missing or stale information. The input is meant to be flexible to allow for different sources of importance. We identify missing facts from a completeness analysis of the coverage of our KG, stale and/or incorrect facts from a human graded fact verification pipeline that uses a traffic-weighted sample of facts from a random sample of anonymous virtual assistant queries, and missing facts of high profile events and social media celebrities. In an assessment of the time lag of facts between Wikidata and Wikipedia, we found on average a 69 day lag of facts from the aforementioned random sample queries and a 85 day lag of facts from social media escalations, with Wikipedia being fresher on average.  Some of the top properties of stale facts are height, population, age, net worth, weight, unmarried partner, child, inception, life expectancy, and date of birth. Our search-based extraction pipeline would use multiple documents from different domains returned by search engines to extract missing and stale facts for these properties. 

While the missing and stale facts will be addressed by the search-based pipeline, we can also initiate extraction in \system by monitoring recent changes from reliable open domain knowledge source, for example, Wikipedia. We setup both daily batch extraction pipeline and hourly streaming extraction pipeline that, at this moment, would monitor changes over English and Spanish Wikipedia pages and trigger \system extraction. 

The Initiator outputs a database containing \triple{$\langle$subject, predicate, url$\rangle$} tuples, which are used as input to the Evidence Retriever component. For example, given a tuple \triple{$\langle$Michael Jordan, Place-of-Birth, \href{https://en.wikipedia.org/wiki/Michael\_Jordan}{en.wikipedia.org/wiki/Michael\_Jordan}$\rangle$}, the Retriever will be tasked to find the birthplace of Michael Jordan using information from his Wikipedia page.

\subsection{Evidence Retriever}
The next step is to retrive documents that are likely to contain supporting evidence for the fact. \system supports two types of retrieval:
\begin{itemize}[itemsep=0pt,parsep=0pt,topsep=0pt,partopsep=0pt,leftmargin=*,labelsep=1mm]
    \item Crawl-based: here, we assume the domain is predefined (for example, Wikipedia), and the mapping from the subject entity to the URL(s) is deterministic (for example, Barack Obama entity can be mapped to his Wikipedia page in multiple languages). Thus, retrieving the evidence is equivalent to simply retrieving pages from a Web Crawl Index.
    Also, in the streaming use case, the url of the changed web document (e.g., a specific Wikipedia page) will sent to the retriever, which will simply retrieve the newly crawled page for downstream extraction. 

    \smallskip
    \item Search-based: here, we assume that either the domain is not known, or the mapping from the subject to the specific URL(s) is not clear: as such, we need to use a search-based system to find relevant web pages. Based on the properties of the missing / stale fact (such as, subject's entity type, subject's name/aliases, and the predicate), we generate one or more search queries using query templates, and send them to our internal Web Search Engine, to retrieve and crawl the URLs that are likely to contain fact evidence.
\end{itemize}



\subsection{Knowledge Extractor}\label{arch:extractor}
Given the retrieved evidence document, and a list of facts to be extracted, the Extractor component's job is to extract the fact's value (e.g. a date for \predicate{Date-of-Birth} predicate, a dollar amount for \predicate{Networth} predicate, or a link to a place entity in \predicate{Place-of-birth} predicate) and provenance map this value to a specific span on a web page. \system employs a variety of extractors, roughly subdivided into \textbf{pattern-based} (using high-precision, domain-specific patterns) and \textbf{model-based} (using trained and zero-shot information extraction models). Below we describe each in detail.

\paragraph{Pattern-based}

Pattern-based extractors are designed to extract simple relation instances of the form $(s, r, o)$ for semi-structured data sources, such as infoboxes from Wikipedia (e.g., \triple{$\langle$Joe Biden, Date-of-Birth, 1942-11-20$\rangle$}). For simplicity, each infobox is treated as an independent list of key-value pairs, where each row constitutes a pair. We write \emph{extraction rules} for rows of infoboxes that converts each row to multiple facts. An extraction rule contains the following components:
\begin{itemize}
    \item {\bf Predicate mapper:} a deterministic function that maps the key to a KG relation. Taking \autoref{fig:example}(a) as an example, the \emph{Height} infobox key could be mapped to the KG relation \texttt{P2048} (height) under the ontology of Wikidata;
    \smallskip
    \item {\bf Value extractor:} Regular expressions to extract values from the infobox table cell. Note that sometimes more than one value could be extracted. Using the same example, a metric height regex is able to extract the value \texttt{184 cm}, whereas an imperial height regex  is able to extract the value \texttt{6 ft 0 in}. If the value contains a hyperlink text span that directly linked to another entity, that entity is directly extracted as the value.
    \smallskip
    \item {\bf Value aggregator:} Given the potentially multiple values extracted in the previous step, we apply an \emph{aggregator} to summarize the values into a most accurate score while removing outliers.
\end{itemize}

We design a flexible framework for defining such extraction rules, with a type validator to check if the rules obey the type constraints (e.g. the object of \texttt{P19} (\textsc{place-of-birth}) has to be an instance of type \texttt{Q2221906} (geographic location)) or other similar types of the KG ontology. More importantly, the framework supports multiple languages. Developers just need to develop language-specific patterns, which can be easily added to the framework. For infoboxes particularly, we have a special type of extractors called link extractor, which would extract facts in infoboxes with hyperlinks. Note that if a fact has a hyperlink, it means itself is a Wikipedia entities, in this case, we would extract the Wikipedia ID (or the equivalent Wikidata QID). Therefore, the link extractor is language-agnostic, and it is frequently used in both our English and Spanish Wikipedia extraction pipelines. 

\paragraph{Model-based}
here, we refer to model-based extractors as the traditional machine learning-based extractors (in contrast to the large language model-based extractors). In this case, we employ a trained {machine reading comprehension} (MRC) model for identifying the fact's value, given a passage from the evidence document. 

The goal of MRC \cite{rajpurkar2016squad,zhang2018record,kwiatkowski2019natural} is, given a question and an evidence document, to produce one or few candidate answers. To generate questions, \system uses the stale / missing fact and one or few question templates: e.g. an incomplete fact \triple{$\langle$Barack Obama, Date-of-Birth, ?$\rangle$} can be converted to ``\textit{When was Barack Obama born?}'' or ``\textit{What is the date of birth of Barack Obama?}'' questions. The \system MRC extractor is based on DeBERTa \cite{he2021deberta} and TaPas \cite{herzig-etal-2020-tapas}, to support extraction from plain text and web tables, respectively. MRC models are mostly used in the search-based extraction, i.e., when we need to find the answer from a few text snippets returned by web search engines. 

\system also supports knowledge extraction by utilizing Question Answering (QA) style LLM prompting \cite{brown2020language}: given an entity and their corresponding evidence document, e.g., entity's Wikipedia page,  \system uses one or few prompts to pose questions about all missing / stale facts of a given entity. When the number of missing / stale facts is large, this approach can be more efficient than MRC, which needs to issue one or few queries \textit{per fact}. The \system LLM extractor is built in a model-agnostic way and was tested with a number of LLM models. Note that, LLM-based extractors are computationally expensive so we only apply LLM-based extractors when the extraction tasks are ambiguous and challenging.

\subsection{Corroborator}\label{arch:corr}
Once the missing / stale facts are extracted in a textual format, \system will normalize the textual facts into standardized forms according to their types; for example, \system employs open-source Duckling \cite{Duckling} to normalize numeric expressions (e.g., dates, heights) and an entity linking model that links the textual span to an existing entity in the KG. Once the candidate facts are extracted and normalized, \system uses the Corroborator to rank the answers by their trustworthiness. We use features such as the extractor type, the score of the extractor, and the number of occurrences of the answer to deliver the final score and ranking. Similar or identical answers are aggregated and scored at this phase using heuristic approaches and/or learning-based algorithms.  In heuristic scoring, extracted normalized facts will be reranked using predefined rules applied to their supporting pieces of evidence. \system also supports model-based ranking with AutoML packages such as H2O \cite{H2OAutoML20} for more complicated fact ranking scenarios. 

\ignore{\subsubsection*{Heuristic approaches}
In heuristic scoring, extracted normalized facts will be reranked using predefined rules applied to their supporting pieces of evidence.

\subsubsection*{Learning-based algorithms}
For learning-based fact ranking, we employ the open-source H2O Auto-ML package \cite{H2OAutoML20}, a unified, scalable,  and explainable framework for parameter tuning and comparison of various models. AutoML allows us to focus on the data and the business problems. As a result of employing AutoML, and compared to ensemble learning methods such as random forest, the performance of the fact-scoring component gained overall $~X\%$ in terms of coverage and accuracy. }

\subsection{Data Export \& KG Ingestion}\label{subsec:ingestion}
The data exported from the exporter in the previous is in \triple{$\langle$s,p,o$\rangle$} triple
format.  This is treated as one of the sources of the KG.  The exported
data then goes through the source ingestion process, where it is
normalized to common ontology. \ignore{After this step all the sources have 
their data is a common ontology and format that allows further processing steps
that can happen.} In cases where the triple is
extracted from Wikipedia or any source that has global unique ID, like
Wikidata QID, we use this ID to link the extracted triple with an
existing entity in the KG.  In cases, where the data is extracted
from the Web, it is likely  there is no such global ID present in the data
or that there is no existing entity to link the triple in the KG. In
these cases,  we run ML models to determine if  there is an existing
entity to link to. If no such entity is found a new entity is created
and added to the graph.  
\ignore{\todo{add figure}}

\subsection{Human Curation}
While a large number of facts can be extracted automatically, there are still ambiguous and sensitive cases that need to be verified by human. Human curation is seamlessly integrated with \system framework. When involve human curation, for example, to extract correct YouTube channel ID for music artists, the curators are first presented with the artist’s entity and properties from the knowledge graph, then asked to verify if the auto-extracted YouTube channel refers to the correct entity. During this evaluation, multiple facets of the artist, including images, alias, short abstract, list of albums can be used to verify the match. The curation task is auto generated based on confidence of the extractor and the complexity of the facts. 

\subsection{Link Inference}
\label{sec:linkinference}
Link inference is orthogonal to the extraction pipeline, which can improve the completeness and correctness of KG by inferring additional edges (facts) without actual extraction. We designed a generic inference pipeline that is running based on the configurations where a user can specify different patterns for different use cases.

\begin{itemize}
    \item{\bf Completeness:} we infer both symmetrical and asymmetrical relationships. For examples, for symmetrical relationship, if we have \spotriple{A,has\_spouse,B} triple in KG, we can infer \spotriple{B,has\_spouse,A} triple if it does not exist in KG. For asymmetrical relationship, if we have \spotriple{A,has\_child,B} triple in our graph, we can infer \spotriple{B,has\_father or has\_mother,A} triple depending on the gender of A.
    \smallskip
    \item {\bf Correctness:} We run the link inference over the high-confidence facts in KG to generate the reverse linkages. We check if the new facts have conflicts with existing facts in our graph. If so, we will correct them. For example, we can inference \spotriple{city\_A,located\_in,county\_B} triple from  \spotriple{county\_B,contain\_cities,city\_A} triple with higher confidence score. If we find existing triple \spotriple{city\_A,located\_in,county\_C} in KG, we can correct it with the inferred triple.
\end{itemize}

\subsection{Deployment and Continuous Update}
\label{sec:streaming}
\system is deployed in two modes, \textbf{batch} and \textbf{streaming}, each  providing different SLAs (service-level agreements) to the downstream customers. The batch mode guarantees that all missing / stale facts will be identified and updated at least weekly. The streaming mode focuses on more frequent updates (e.g. hourly) for a small number (e.g. top-1\%) of the most important facts, such as, when a new president of a country has been elected, or when a major sport event has been announced. We are testing the end-to-end runtime of various streaming extraction settings, and we are now able to achieve an SLA of 4 hours starting from a change (no vandalism) being made in Wikipedia ending at the facts being extracted and ingested into KG. 

Once an extraction pipeline is finished, it outputs its extracted facts into a centralized, append-only, versioned fact table. This table is periodically materialized into a view with ``latest'' extracted facts. Then, depending on the mode, \system sends the relevant latest facts to the Ingestion component (\ref{subsec:ingestion}), either via another table (in batch mode) or via a messaging queue (in streaming mode), to make sure the newly extracted facts are added to the KG in a timely manner.


\section{Results and Impact}
\system has been successfully deployed to extract and ingest tens of millions of high-quality facts into an industrial KG. In this section, we provide selected evaluation results of \system in real-world applications.  Our evaluation covers both intrinsic and extrinsic metrics. Intrinsic metrics (i.e., evaluate \system's inherent strengths) include:
\begin{itemize}[itemsep=0pt,parsep=0pt,topsep=0pt,partopsep=0pt,leftmargin=*,labelsep=1mm]
    \item \textit{Number of fresher and fresh new facts}: how many fresher and fresh new facts added to KG.
    \smallskip
    \item \textit{Extraction Precision}: out of all extracted facts, how many of them were faithfully. 
    \smallskip
    \item \textit{Throughput}: number of ingestion-ready facts per hour.
\end{itemize}
Extrinsic metrics (i.e., evaluate \system's performance in practical tasks) including \textit{potential product impact} and \textit{User experience improvement}.
To illustrate the effectiveness of \system, we discuss its practical value with three real-world scenarios.

\vspace{1mm}
\subsection{Bridge the freshness gap between Wikidata and Wikipedia}
Wikidata and Wikipedia are two popular open-domain knowledge platforms. While one might expect data consistency between the two, unfortunately, that is not always the case. \ignore{In Fig. \ref{fig:example} (b) and (d), contradictory height values for the same entity can be observed.}
As we mentioned in Section \ref{sec:initiator}, we observed the freshness gap between Wikidata and Wikipedia. To bridge this gap, we developed a pipeline to extract facts from Wikipedia and compare them with corresponding facts in Wikidata.

For this task, we focused on some popular properties of a selected set of entity types (e.g., Date-of-Birth and height for Person entities, screenwriters and producers for Movie entities). As a result, for 19 properties of 9 entity types, \system extracted \textbf{1,059,876} different facts from Wikipedia compared to Wikidata counterparts, consisting of \textbf{563,945} fresher facts (overlapped with Wikidata but different) and \textbf{495,931} fresh new facts. The extraction precision of \systemWikipedia pipeline is \textbf{99.2\%} aggregate across all predicates based on human annotation labels. Table \ref{tab:odke-vs-Wikidata} shows some selected results of the comparison between \systemWikipedia and Wikidata.

\begin{table}
\centering
\normalsize
\renewcommand{\arraystretch}{1.2}
\begin{tabular}{@{}ccccc@{}}
\textbf{Type}           & \textbf{Property} & \textbf{\begin{tabular}[c]{@{}c@{}}\system\\delta \%\end{tabular}} & \textbf{\# Fresher} & \textbf{\# Fresh} \\
\specialrule{\heavyrulewidth}{0pt}{\belowrulesep}
\multirow{2}{*}{Person} & Height            & 33.6\%                                                                             & $\sim$254k          & $\sim$340k        \\
\cmidrule(lr){2-5} 

                        & Birthplace        & 13.2\%                                                                             & $\sim$387k          & $\sim$49k         \\
\hline
\multirow{2}{*}{Movie}  & Producer          & 10.9\%                                                                             & 6464                & 15307             \\\cmidrule(lr){2-5}
                        & Screenwriter      & 8.6\%                                                                              & 13408               & 17810 \\
\specialrule{\heavyrulewidth}{0pt}{\belowrulesep}
\end{tabular}
\caption{\systemWikipedia vs. Wikidata for selected entity-property groups}
\label{tab:odke-vs-Wikidata}
\end{table}

While the difference between \system-Wikipedia and Wikidata is substantial, it is crucial to understand if this difference translates into an improved user experience. To address it, we sampled disagreements between \system-Wikipedia and Wikidata, focusing on two downstream practical use cases. We presented these discrepancies side-by-side to human annotators and asked them to provide their preference between the two options without revealing the source of the data.
\begin{table}[h!]
\centering
\normalsize
\renewcommand{\arraystretch}{1.0}
\begin{tabular}{@{}cccc@{}}
\multicolumn{1}{l}{}                                                & \textbf{\begin{tabular}[c]{@{}c@{}}\system is\\ better\end{tabular}} & \textbf{\begin{tabular}[c]{@{}c@{}}Equally\\ good\end{tabular}} & \textbf{\begin{tabular}[c]{@{}c@{}}Wikidata is\\ better\end{tabular}} \\ \specialrule{\heavyrulewidth}{0pt}{\belowrulesep}
\begin{tabular}[c]{@{}c@{}}Use case I\\ (138 samples)\end{tabular}  & \textbf{130 (94\%)}                                                  & 7                                                               & 0                                                                     \\
\cmidrule(lr){2-4} 
\begin{tabular}[c]{@{}c@{}}Use case II\\ (211 samples)\end{tabular} & \textbf{193 (91.5\%)}  & 19 & 0
\\
\specialrule{\heavyrulewidth}{0pt}{\belowrulesep}
\end{tabular}
\caption{Human Preference: \system vs. Wikidata}
\label{tab:odke-Wikipedia-quality}
\end{table}
As shown in Table \ref{tab:odke-Wikipedia-quality}, the majority ($>$90\%) of facts extracted by \system are preferred over Wikidata facts by human judges. This clear preference highlights the effectiveness of the \systemWikipedia pipeline.

\ignorebecauseofproduct{Finally, we also estimated the product impact of the delta portion of \system-Wikipedia based on a sample of historical traffic. Our projection suggests that the weighted traffic impact of this work is \textbf{1.6\%}. Considering the volume of our end customers, this impact is significant. }

\vspace{2mm}
\subsection{New Trendy Entity Discovery}
As mentioned earlier, new entities such as B-list celebrities emerge everyday. To provide their up-to-date information with their fans, \system was used to build new trendy entity discovery and ingestion pipeline to power this application. \system was used to build pipelines for two popular online data platforms dedicated to provide reliable information for notable people from various fields, including actors, musical artists, athletes, social media influencers etc, which we refer to as ``\fandom'' and ``\famousbirthday''. 

We focused on a subset of the Person domain in the two platforms and conducted crawl-based extraction. We extracted Person entities including all their available key facts (e.g.,  occupation, date of birth (DoB), birthplace, blood type, and social media) from both platforms. 
\begin{table}[h]
\centering
\normalsize
\renewcommand{\arraystretch}{1}
\begin{tabular}{@{}cccc@{}}
& \textbf{\begin{tabular}[c]{@{}c@{}}\# Entities \\ Extracted\end{tabular}} & \textbf{\begin{tabular}[c]{@{}c@{}}\# Prop.\\ covered\end{tabular}} & \textbf{\begin{tabular}[c]{@{}c@{}}\# New \\ Entities\end{tabular}} \\ \specialrule{\heavyrulewidth}{0pt}{\belowrulesep}
\fandom         & 5,239                                                                     & 10                                                                       & 1,678 (32.3\%)                                                      \\ \cmidrule(lr){2-4} 
\famousbirthday & 249,858                                                                   & 25                                                                       & 67,609 (27.05\%)                                                    \\ \specialrule{\heavyrulewidth}{0pt}{\belowrulesep}

\end{tabular}
\caption{New Entity Discovery Statistics (Prop. is the abbreviation of Properties)}
\label{tab:new-entity-stats}
\end{table}
Table \ref{tab:new-entity-stats} shows the total number of entities \system extracted from the two platforms, number of properties covered in the extraction, and how many of these extracted entities resulted in new entities after entity disambiguation with KG. The extraction precision were all above \textbf{95\%} with most of them being \textbf{100\%}. 

\vspace{2mm}
\subsection{Add Missing Facts through Link inference}
We have performed the link inference on certain properties and observed meaningful results. Table \ref{tab:new-entity-link-infer} shows the number of new facts inferred. 

\begin{table}[h]
\centering
\normalsize
\begin{tabular}{cc}
\textbf{\begin{tabular}[c]{@{}c@{}} Property \end{tabular}}  & \textbf{\begin{tabular}[c]{@{}c@{}}\# New Facts\end{tabular}} \\ \specialrule{\heavyrulewidth}{0pt}{\belowrulesep}
has\_child         & 27,403                                                                      \\ 
has\_mother        & 6,329                                                                       \\ 
has\_father        & 13,795                                                                      \\ 
has\_spouse        & 41,108                                                                       \\ 
\specialrule{\heavyrulewidth}{0pt}{\belowrulesep}
\end{tabular}
\caption{New Facts through Link Inference}
\label{tab:new-entity-link-infer}
\end{table}
The newly inferred facts were evaluated by human, and it shows that the accuracy is 99.7\%. We are actively exploring the potential property candidates that we can run the link inference on.

\subsection{Throughput}
High automation and scalability are two major benefits offered by \system. As mentioned earlier, \system supports both batch mode and streaming mode. The ODKE-Wikipedia pipeline is scheduled for both weekly mode and streaming mode, where the batch mode would process about 6.7 million English Wikipedia pages weekly. End-to-end, each run of the pattern-based ODKE-Wikipedia extraction pipeline takes averagely 69 mins to finish and can extract up to 6 million facts per hour, or up to 100K facts per minute. This gives us up to 4000x speedup compared to an existing human-based curation pipeline. 
We also scheduled hourly run of ODKE-Wikipedia pipeline, where we will monitor and collect the recent Wikipedia edits (after removing potential vandalism), and trigger our extraction pipelines every hour. The output of the hourly streaming pipeline will be ingested into KG to provide fresh knowledge for downstream applications. 

We are also testing the weekly batch pipelines with LLM-based extractors. With very large models with 60-70 billion parameters, LLM-based pipelines can process up to 200K Wikipedia documents per day with our batch scrapers. Therefore, LLM-based pipelines are only enabled for extraction tasks that pattern-based extractors cannot perform well. However, LLM-based extractors can be used in hourly streaming pipelines given that the hourly delta changes (e.g., a few hundred Wikipedia pages) is relatively small. 

\ignorebecauseofproduct{leading to about 500,000 USD cost saving when performing a task of 1 million extraction jobs\footnote{All these numbers are estimated based on our internal cost caculation}. }

\section{Concluding Remarks}
\system offers a comprehensive and automated solution for knowledge extraction and ingestion at an industrial scale. With a powerful and carefully designed deployment plan, \system enables efficient extraction and ingestion of high-quality facts, seamlessly supporting near real-time streaming and offline batch pipelines. \ignorebecauseofproduct{Its successful deployment in production, ingesting millions of facts into an industrial knowledge graph, demonstrates its effectiveness in powering products for a large number of customers.}

We also would like to mention that our research respects user privacy and ensures no misuse of personal information. Additionally, while LLMs can be integrated with \system pipelines to extract facts, the assets of LLM-based extractors are still used for evaluation purposes only, rather than being ingested into the KG to support real-world features. This approach is taken due to potential bias concerns associated with LLMs.

\bibliographystyle{IEEEtran}
\bibliography{odke, icdebib}

\end{document}